\documentclass[10pt,twocolumn,letterpaper]{article}

\usepackage{cvpr}
\usepackage{times}
\usepackage{epsfig}
\usepackage{graphicx}
\usepackage{amsmath}
\usepackage{amssymb}
\usepackage{mathtools}
\usepackage{verbatim}

\usepackage{booktabs} %
\usepackage{multirow}
\usepackage{array}
\usepackage{footnote}

\usepackage[linesnumbered,lined,vlined,ruled,commentsnumbered]{algorithm2e}
\usepackage[pagebackref=true,breaklinks=true,letterpaper=true,colorlinks,bookmarks=false]{hyperref}
\cvprfinalcopy %

\newcommand{\beginsupp}{%
        \setcounter{table}{0}
        \renewcommand{\thetable}{S\arabic{table}}%
        \setcounter{figure}{0}
        \renewcommand{\thefigure}{S\arabic{figure}}%
     }
\newcommand{\myparagraph}[1]{\vspace{0.1em}\noindent\textbf{#1}}

\newcommand{\redt}[1]{\textcolor[rgb]{1,0,0}{#1}}
\begin{document}

\title{Meta-Transfer Learning for Few-Shot Learning}
\author{Qianru Sun$^{1,3}$ \quad Yaoyao Liu$^{2}$ \quad Tat-Seng Chua$^{1}$ \quad Bernt Schiele$^{3}$ \\
\\
\small  $^{1}$National University of Singapore \quad  $^{2}$Tianjin University$\thanks{Yaoyao Liu did this work during his internship at NUS.}$ \\
\small  $^{3}$Max Planck Institute for Informatics, Saarland Informatics Campus\\
\small  {\texttt{\{qsun, schiele\}@mpi-inf.mpg.de}} \\
\small  {\texttt{liuyaoyao@tju.edu.cn}} \quad  {\texttt{\{dcssq, dcscts\}@nus.edu.sg}}
}

\maketitle
\thispagestyle{empty}

\begin{abstract}
Meta-learning has been proposed as a framework to address the challenging few-shot learning setting. The key idea is to leverage a large number of similar few-shot tasks in order to learn how to adapt a base-learner to a new task for which only a few labeled samples are available. As deep neural networks (DNNs) tend to overfit using a few samples only, meta-learning typically uses shallow neural networks (SNNs), thus limiting its effectiveness. 
In this paper we propose a novel few-shot learning method called \textbf{meta-transfer learning (MTL)} which learns to adapt a \textbf{deep NN} for \textbf{few shot learning tasks}. Specifically, \emph{meta} refers to training multiple tasks, and \emph{transfer} is achieved by learning scaling and shifting functions of DNN weights for each task.
In addition, we introduce the \textbf{hard task (HT) meta-batch} scheme as an effective learning curriculum for MTL. We conduct experiments using (5-class, 1-shot) and (5-class, 5-shot) recognition tasks on two challenging few-shot learning benchmarks: miniImageNet and Fewshot-CIFAR100. Extensive comparisons to related works validate that our \textbf{meta-transfer learning} approach trained with the proposed \textbf{HT meta-batch} scheme achieves top performance. An ablation study also shows that both components contribute to fast convergence and high accuracy\footnote{Code: \href{https://github.com/y2l/meta-transfer-learning-tensorflow}{https://github.com/y2l/meta-transfer-learning-tensorflow}}.


\end{abstract}

\section{Introduction}

While deep learning systems have achieved great performance when sufficient amounts of labeled data are available~\cite{Lecun2015, HeZRS16, ShelhamerLD17},
there has been growing interest in reducing the required amount of data.
Few-shot learning tasks have been defined for this purpose. 
The aim is to learn new concepts from few labeled examples, e.g. 1-shot learning~\cite{FeiFeiFP06}.
While humans tend to be highly effective in this context, often grasping the essential connection between new concepts and their own knowledge and experience, it remains challenging for machine learning approaches.
E.g., on the CIFAR-100 dataset, a state-of-the-art method~\cite{OreshkinNIPS18} achieves only $40.1\%$ accuracy for 1-shot learning, compared to $75.7\%$ for the all-class fully supervised case~\cite{ClevertUH15}.

\begin{figure}[t]
  \centering
  \includegraphics[width=0.99\linewidth]{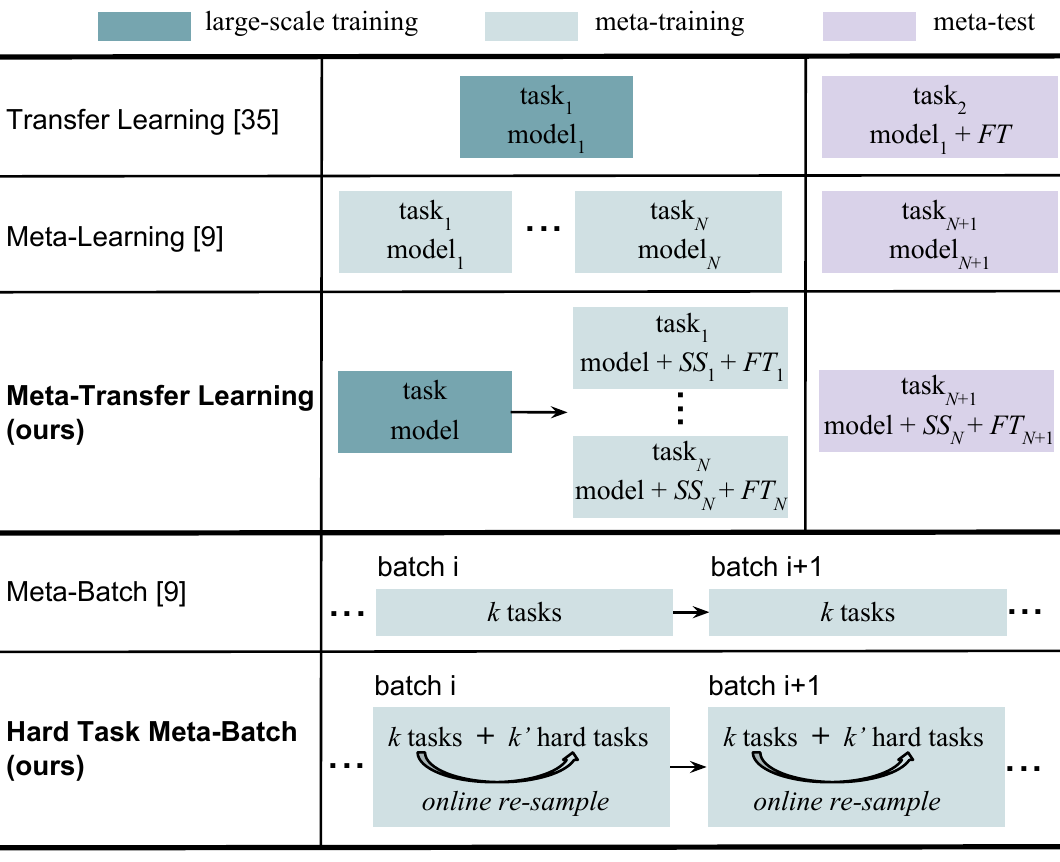}
     \caption{Meta-transfer learning (MTL) is our meta-learning paradigm and hard task (HT) meta-batch is our training strategy. The upper three rows show the differences between MTL and related methods, transfer-learning~\cite{PanTKY11} and meta-learning~\cite{FinnAL17}.
     The bottom rows compare HT meta-batch with the conventional meta-batch~\cite{FinnAL17}. \emph{FT} stands for fine-tuning a classifier. \emph{SS} represents the \emph{Scaling} and \emph{Shifting} operations in our MTL method.}
  \label{main_framework_tisser}
  \vspace{-0.3cm}
\end{figure}

Few-shot learning methods can be roughly categorized into two classes: data augmentation and task-based meta-learning.
Data augmentation is a classic technique to increase the amount of available data and thus also useful for few-shot learning~\cite{KhorevaBIBS17}.
Several methods propose to learn a data generator e.g. conditioned on Gaussian noise~\cite{Mehrotra2017, SchwartzNIPS18, WangCVPR2018}. 
However, the generation models often underperform when trained on few-shot data~\cite{BartunovV18}.
An alternative is to merge data from multiple tasks which, however, is not effective due to variances of the data across tasks~\cite{WangCVPR2018}.

In contrast to data-augmentation methods, meta-learning is a task-level learning method~\cite{Bengio92, Naik92, Thrun1998}.
Meta-learning aims to accumulate experience from learning multiple tasks \cite{FinnAL17, RaviICLR2017, SnellSZ17, MunkhdalaiICML2017, GrantICLR2018}, while base-learning focuses on modeling the data distribution of a single task. 
A state-of-the-art representative of this, namely 
Model-Agnostic Meta-Learning (MAML), learns to search for the optimal initialization state to fast adapt a base-learner to a new task~\cite{FinnAL17}.
Its task-agnostic property makes it possible to generalize to few-shot supervised learning as well as unsupervised reinforcement learning~\cite{GrantICLR2018, FinnNIPS2018}.
However, in our view, there are two main limitations of this type of approaches limiting their effectiveness: i)~these methods usually require a large number of similar tasks for meta-training which is costly; and ii)~each task is typically modeled by a low-complexity base learner (such as a shallow neural network) to avoid model overfitting, thus being unable to use deeper and more powerful architectures. 
For example, for the miniImageNet dataset~\cite{VinyalsBLKW16}, MAML uses a \emph{shallow} CNN with only $4$ CONV layers and its optimal performance was obtained learning on $240k$ tasks.

In this paper, we propose a novel meta-learning method called \textbf{meta-transfer learning (MTL)} leveraging the advantages of both transfer and meta learning (see conceptual comparison of related methods in Figure~\ref{main_framework_tisser}). 
In a nutshell, MTL is a novel learning method that helps deep neural nets converge faster while reducing the probability to  overfit when using few labeled training data only.  
In particular, ``transfer'' means that DNN weights trained on large-scale data can be used in other tasks by two light-weight neuron operations: \emph{Scaling} and \emph{Shifting} (\emph{SS}), i.e. $\alpha X+\beta$. ``Meta'' means that the parameters of these operations can be viewed as hyper-parameters trained on few-shot learning tasks~\cite{MunkhdalaiICML2017, LiICML2018}.
Large-scale trained DNN weights offer a good initialization, enabling fast convergence of meta-transfer learning with fewer tasks, e.g. only $8k$ tasks for miniImageNet~\cite{VinyalsBLKW16}, $30$ times fewer than MAML~\cite{FinnAL17}.
Light-weight operations on DNN neurons have less parameters to learn, e.g. less than $\tfrac{2}{49}$ if considering neurons of size $7\times 7$ ($\tfrac{1}{49}$ for $\alpha$ and $<\tfrac{1}{49}$ for $\beta$), reducing the chance of overfitting.
In addition, these operations keep those trained DNN weights unchanged, and thus avoid the problem of ``catastrophic forgetting'' which means forgetting general patterns when adapting to a specific task~\cite{LopezPazNIPS17, McCloskey1989}.

The second main contribution of this paper is an effective meta-training curriculum.
Curriculum learning~\cite{BengioLCW09} and hard negative mining~\cite{ShrivastavaGG16} both suggest that faster convergence and stronger performance can be achieved by a better arrangement of training data.
Inspired by these ideas, we design our \textbf{hard task (HT) meta-batch} strategy to offer a challenging but effective learning curriculum.
As shown in the bottom rows of Figure~\ref{main_framework_tisser}, a conventional meta-batch contains a number of random tasks~\cite{FinnAL17}, but our HT meta-batch online re-samples harder ones according to past failure tasks with lowest validation accuracy.

Our overall contribution is thus three-fold:
i)~we propose a novel \textbf{MTL} method that learns to transfer large-scale pre-trained DNN weights for solving few-shot learning tasks; 
ii)~we propose a novel \textbf{HT meta-batch} learning strategy that forces meta-transfer to ``grow faster and stronger through hardship''; and 
iii)~we conduct extensive experiments on two few-shot learning benchmarks, namely miniImageNet~\cite{VinyalsBLKW16} and Fewshot-CIFAR100 (FC100)~\cite{OreshkinNIPS18}, and achieve the state-of-the-art performance. 

\section{Related work}
\myparagraph{Few-shot learning}
Research literature on few-shot learning exhibits great diversity. 
In this section, we focus on methods using the supervised meta-learning paradigm~\cite{Hinton1987, Thrun1998, FinnAL17} most relevant to ours
and compared to in the experiments. 
We can divide these methods into three categories.
1)~\emph{Metric learning} methods \cite{VinyalsBLKW16, SnellSZ17, SungCVPR2018} learn a similarity space in which learning is efficient for few-shot examples.
2)~\emph{Memory network} methods \cite{MunkhdalaiICML2017, SantoroBBWL16, OreshkinNIPS18, MishraICLR2018} learn to store ``experience'' when learning seen tasks and then generalize that to unseen tasks.
3)~\emph{Gradient descent} based methods \cite{FinnAL17, RaviICLR2017, LeeICML18, GrantICLR2018, ZhangNIPS2018MetaGAN} have a specific \emph{meta-learner} that learns to adapt a specific \emph{base-learner} (to few-shot examples) through different tasks. E.g. MAML \cite{FinnAL17} uses a meta-learner that learns to effectively initialize a base-learner for a new learning task.
Meta-learner optimization is done by gradient descent using the validation loss of the base-learner.
Our method is closely related.
An important difference is that our MTL approach leverages transfer learning and benefits from referencing neuron knowledge in pre-trained deep nets. Although MAML can start from a pre-trained network, its element-wise fine-tuning makes it hard to learn deep nets without overfitting (validated in our experiments).

\begin{figure*}[htp]
  \centering
  \includegraphics[width=0.99\linewidth]{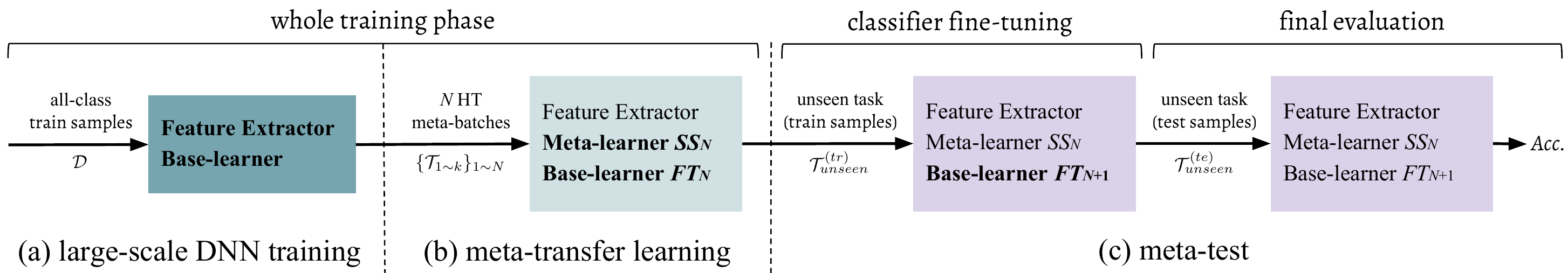}
     \caption{The pipeline of our proposed few-shot learning method, including three phases: (a) DNN training on large-scale data, i.e. using all training datapoints (Section~\ref{sec_large_scale_pretrain}); (b) Meta-transfer learning (MTL) that learns the parameters of \emph{Scaling} and \emph{Shifting} (\emph{SS}), based on the pre-trained feature extractor (Section~\ref{sec_meta_transfer}). Learning is scheduled by the proposed HT meta-batch (Section~\ref{sec_HT}); and (c) meta-test is done for an unseen task which consists of a base-learner (classifier) \emph{Fine-Tuning} (\emph{FT}) stage and a final evaluation stage, described in the last paragraph in Section~\ref{sec_preli}. 
     Input data are along with arrows. Modules with names in bold get updated at corresponding phases. Specifically, \emph{SS} parameters are learned by meta-training but fixed during meta-test. Base-learner parameters are optimized for every task.}
     \vspace{-0.3cm}
  \label{main_framework_meta_transf_hard_task}
\end{figure*}

\myparagraph{Transfer learning}
\emph{What} and \emph{how} to transfer are key issues to be addressed in transfer learning, as different methods are applied to different source-target domains and bridge different transfer knowledge~\cite{PanTKY11, YangICDM07, WeiICML2018, AmirCVPR18}.
For deep models, a powerful transfer method is adapting a pre-trained model for a new task, often called \emph{fine-tuning} (\emph{FT}). Models pre-trained on large-scale datasets have proven to generalize better than randomly initialized ones \cite{Erhan10}.
Another popular transfer method is taking pre-trained networks as backbone and adding high-level functions, e.g. for object detection and recognition~\cite{HuangCVPR017, Sun_2017_CVPR, Sun_2018_CVPR} and image segmentation~\cite{He_MaskRCNN17,ChenPAMI18}.
Our meta-transfer learning leverages the idea of transferring
pre-trained weights and aims to meta-learn how to effectively transfer. 
In this paper, large-scale trained DNN weights are \emph{what} to transfer, and the operations of \emph{Scaling} and \emph{Shifting} indicate \emph{how} to transfer.
Similar operations have been used to modulating the per-feature-map distribution of activations for visual reasoning~\cite{FiLM2018}.

Some few-shot learning methods have been proposed to use pre-trained weights as initialization \cite{Keshari18, MishraICLR2018, QiaoCVPR2018, ScottNIPS2018, Rusu2019}. Typically, weights are fine-tuned for each task, while we learn a meta-transfer learner through all tasks, which is different in terms of the underlying learning paradigm. 

\myparagraph{Curriculum learning \& Hard sample mining}
Curriculum learning was proposed by Bengio \emph{et al}. \cite{BengioLCW09} and is popular for multi-task learning~\cite{PentinaCVPR15, SarafianosGNK17, WeinshallCA18, GravesICML2017}.
They showed that instead of observing samples at random it is better to organize samples in a meaningful way so that fast convergence, effective learning and better generalization can be achieved.
Pentina \emph{et al}. \cite{PentinaCVPR15} use adaptive SVM classifiers to evaluate task difficulty for later organization. Differently, our MTL method does task evaluation online at the phase of episode test, without needing any auxiliary model.

Hard sample mining was proposed by Shrivastava \emph{et al}.~\cite{ShrivastavaGG16} for object detection. It treats image proposals overlapped with ground truth as hard negative samples. Training on more confusing data enables the model to achieve higher robustness and better performance~\cite{CanevetF16, HarwoodGCRD17, DalalT05}. 
Inspired by this, we sample harder tasks online and make our MTL learner ``grow faster and stronger through more hardness''. In our experiments, we show that this can be generalized to enhance other meta-learning methods, e.g. MAML~\cite{FinnAL17}.

\section{Preliminary}
\label{sec_preli}

We introduce the problem setup and notations of meta-learning, following related work~\cite{VinyalsBLKW16, RaviICLR2017, FinnAL17, OreshkinNIPS18}.

\myparagraph{Meta-learning} consists of two phases: meta-train and meta-test. A meta-training example is a classification task $\mathcal{T}$ sampled from a distribution $p(\mathcal{T})$. 
$\mathcal{T}$ is called episode, including a training split $\mathcal{T}^{(tr)}$ to optimize the base-learner, and a test split $\mathcal{T}^{(te)}$ to optimize the meta-learner. 
In particular, meta-training aims to learn from a number of episodes $\{\mathcal{T}\}$ sampled from $p(\mathcal{T})$.
An unseen task $\mathcal{T}_{unseen}$ in meta-test will start from that experience of the meta-learner and adapt the base-learner. The final evaluation is done by testing a set of unseen datapoints $\mathcal{T}^{(te)}_{unseen}$.

\myparagraph{Meta-training phase.} 
This phase aims to learn a meta-learner from multiple episodes.
In each episode, meta-training has a two-stage optimization. 
Stage-1 is called base-learning, where the cross-entropy loss is used to optimize the parameters of the base-learner.
Stage-2 contains a feed-forward test on episode test datapoints. The test loss is used to optimize the parameters of the meta-learner.
Specifically, given an episode $\mathcal{T} \in p(\mathcal{T})$, the base-learner $\theta_\mathcal{T}$ is learned from episode training data $\mathcal{T}^{(tr)}$ and its corresponding loss $\mathcal{L}_{\mathcal{T}}(\theta_\mathcal{T}, \mathcal{T}^{(tr)})$. 
After optimizing this loss, the base-learner has parameters $\tilde{\theta}_\mathcal{T}$. 
Then, the meta-learner is updated using test loss $\mathcal{L}_{\mathcal{T}}(\tilde{\theta}_\mathcal{T}, \mathcal{T}^{(te)})$. 
After meta-training on all episodes, the meta-learner is optimized by test losses $\{\mathcal{L}_{\mathcal{T}}(\tilde{\theta}_\mathcal{T}, \mathcal{T}^{(te)})\}_{\mathcal{T} \in p(\mathcal{T})}$. Therefore, the number of meta-learner updates equals to the number of episodes.

\myparagraph{Meta-test phase.} 
This phase aims to test the performance of the trained meta-learner for fast adaptation to unseen task.
Given $\mathcal{T}_{unseen}$, the meta-learner $\tilde{\theta}_\mathcal{T}$ teaches the base-learner $\theta_{\mathcal{T}_{unseen}}$ to adapt to the objective of $\mathcal{T}_{unseen}$ by some means, e.g. through initialization~\cite{FinnAL17}. 
Then, the test result on $\mathcal{T}^{(te)}_{unseen}$ is used to evaluate the meta-learning approach. 
If there are multiple unseen tasks $\{\mathcal{T}_{unseen}\}$, the average result on $\{\mathcal{T}^{(te)}_{unseen}\}$ will be the final evaluation.

\section{Methodology}

As shown in Figure~\ref{main_framework_meta_transf_hard_task},  our method consists of three phases.
First, we train a DNN on large-scale data, e.g. on miniImageNet ($64$-class, $600$-shot)~\cite{VinyalsBLKW16}, and then fix the low-level layers as Feature Extractor (Section~\ref{sec_large_scale_pretrain}). 
Second, in the meta-transfer learning phase, MTL learns the \emph{Scaling} and \emph{Shifting} (\emph{SS}) parameters for the Feature Extractor neurons, enabling fast adaptation to few-shot tasks (Section~\ref{sec_meta_transfer}).
For improved overall learning, we use our HT meta-batch strategy  (Section~\ref{sec_HT}).
The training steps are detailed in Algorithm~\ref{alg_overall} in Section~\ref{sec_alg}.
Finally, the typical meta-test phase is performed, as introduced in Section~\ref{sec_preli}.

\subsection{DNN training on large-scale data}
\label{sec_large_scale_pretrain}

This phase is similar to the classic pre-training stage as, e.g., pre-training on Imagenet for object recognition~\cite{Russakovsky2015}.
Here, we do not consider data/domain adaptation from other datasets, and  pre-train on readily available data of few-shot learning benchmarks, allowing for fair comparison with other few-shot learning methods.
Specifically, for a particular few-shot dataset, we merge all-class data $\mathcal{D}$ for pre-training.
For instance, for miniImageNet~\cite{VinyalsBLKW16}, there are totally $64$ classes in the training split of $\mathcal{D}$ and each class contains $600$ samples used to pre-train a $64$-class classifier.

We first randomly initialize a feature extractor $\Theta$ (e.g. CONV layers in ResNets~\cite{HeZRS16}) and a classifier $\theta$ (e.g. the last FC layer in ResNets~\cite{HeZRS16}), and then optimize them by gradient descent as follows,
\begin{equation}\label{eq_large_scale_update}
 [\Theta; \theta] =: [\Theta; \theta] - \alpha\nabla\mathcal{L}_{\mathcal{D}}\big([\Theta; \theta]\big),
\end{equation}
where $\mathcal{L}$ denotes the following empirical loss,
\begin{equation}\label{eq_large_scale_loss}
    \mathcal{L}_{\mathcal{D}}\big([\Theta; \theta]\big) = \frac{1}{|\mathcal{D}|}\sum_{(x,y)\in \mathcal{D}}l\big(f_{[\Theta; \theta]}(x), y\big),
\end{equation}
e.g. cross-entropy loss, and $\alpha$ denotes the learning rate. 
In this phase, the feature extractor $\Theta$ is learned. It will be frozen in the following meta-training and meta-test phases, as shown in Figure~\ref{main_framework_meta_transf_hard_task}. 
The learned classifier $\theta$ will be discarded, because subsequent few-shot tasks contain different classification objectives, e.g. $5$-class instead of $64$-class classification for miniImageNet~\cite{VinyalsBLKW16}.

\subsection{Meta-transfer learning (MTL)}
\label{sec_meta_transfer}

\begin{figure}[t]
  \centering
  \includegraphics[width=0.99\linewidth]{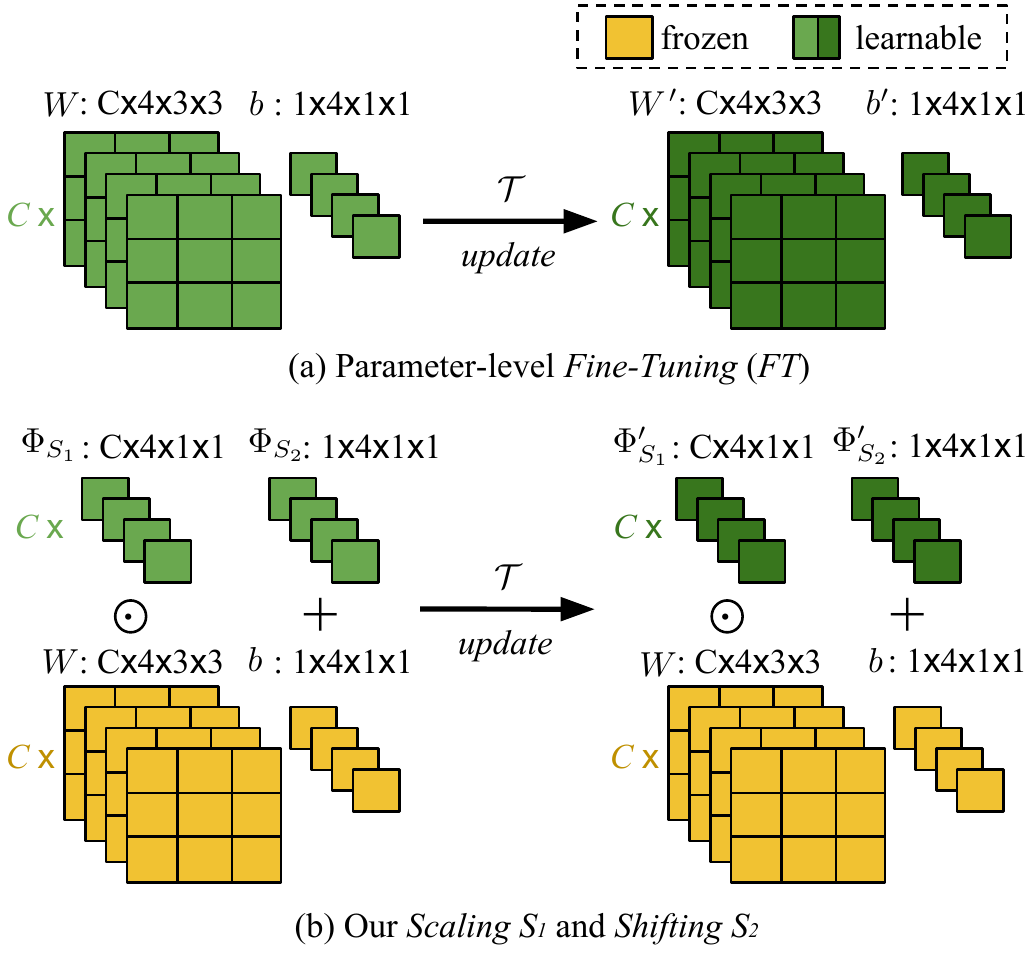}
     \caption{(a) Parameter-level \emph{Fine-Tuning} (\emph{FT}) is a conventional meta-training operation, e.g. in MAML~\cite{FinnAL17}. Its update works for all neuron parameters, $W$ and $b$.
     (b) Our neuron-level \emph{Scaling} and \emph{Shifting} (\emph{SS}) operations in MTL. They reduce the number of learning parameters and avoid overfitting problems. In addition, they keep large-scale trained parameters (in yellow) frozen, preventing ``catastrophic fogetting''~\cite{LopezPazNIPS17, McCloskey1989}.
    }
  \label{main_framework_SS_FT}
\end{figure}

As shown in Figure~\ref{main_framework_meta_transf_hard_task}(b), our proposed meta-transfer learning (MTL) method optimizes the meta operations \emph{Scaling} and \emph{Shifting} (\emph{SS}) through HT meta-batch training (Section~\ref{sec_HT}). 
Figure~\ref{main_framework_SS_FT} visualizes the difference of updating through \emph{SS} and \emph{FT}.
\emph{SS} operations, denoted as $\Phi_{S_1}$ and $\Phi_{S_2}$, do not change the frozen neuron weights of $\Theta$ during learning, while \emph{FT} updates the complete $\Theta$. 

In the following, we detail the \emph{SS} operations.
Given a task $\mathcal{T}$, the loss of $\mathcal{T}^{(tr)}$ is used to optimize the current base-learner (classifier) $\theta'$ by gradient descent:
\begin{equation}\label{eq_base_classifier}
  \theta' \gets \theta - \beta\nabla_{\theta}\mathcal{L}_{\mathcal{T}^{(tr)}}\big([\Theta; \theta], \Phi_{S_{\{1,2\}}}\big),
\end{equation}
which is different to Eq.~\ref{eq_large_scale_update}, as we do not update $\Theta$. 
Note that here $\theta$ is different to the one from the previous phase, the large-scale classifier $\theta$ in Eq.~\ref{eq_large_scale_update}. 
This $\theta$ concerns only a few of classes, e.g. 5 classes, to classify each time in a novel few-shot setting. 
$\theta'$ corresponds to a temporal classifier only working in the current task, initialized by the $\theta$ optimized for the previous task (see Eq.~\ref{eq_meta_classifier}).

$\Phi_{S_1}$ is initialized by ones and $\Phi_{S_1}$ by zeros. Then, they are optimized by the test loss of $\mathcal{T}^{(te)}$ as follows,
\begin{equation}\label{eq_ss_update}
     \Phi_{S_i} =: \Phi_{S_i} - \gamma\nabla_{\Phi_{S_i}}\mathcal{L}_{\mathcal{T}^{(te)}}\big([\Theta; \theta'], \Phi_{S_{\{1,2\}}}\big).
\end{equation}
In this step, $\theta$ is updated with the same learning rate $\gamma$ as in Eq.~\ref{eq_ss_update},
\begin{equation}\label{eq_meta_classifier}
  \theta =: \theta - \gamma\nabla_{\theta}\mathcal{L}_{\mathcal{T}^{(te)}}\big([\Theta; \theta'], \Phi_{S_{\{1,2\}}}\big).
\end{equation}
Re-linking to Eq.~\ref{eq_base_classifier}, we note that the above $\theta'$ comes from the last epoch of base-learning on $\mathcal{T}^{(tr)}$.

Next, we describe how we apply $\Phi_{S_{\{1,2\}}}$ to the frozen neurons as shown in Figure~\ref{main_framework_SS_FT}(b). 
Given the trained $\Theta$, for its $l$-th layer containing $K$ neurons, we have $K$ pairs of parameters, respectively as weight and bias, denoted as $\{(W_{i,k}, b_{i,k})\}$. Note that the neuron location $l, k$ will be omitted for readability.
Based on MTL, we learn $K$ pairs of scalars $\{\Phi_{S_{\{1,2\}}}\}$. 
Assuming $X$ is input, we apply $\{\Phi_{S_{\{1,2\}}}\}$ to $(W, b)$ as
\begin{equation}\label{eq_SS_operation}
    SS(X; W, b; \Phi_{S_{\{1,2\}}}) =(W\odot\Phi_{S_1}) X + (b + \Phi_{S_2}),
\end{equation}
where $\odot$ denotes the element-wise multiplication.

Taking Figure~\ref{main_framework_SS_FT}(b) as an example of a single $3\times 3$ filter, after \emph{SS} operations, this filter is scaled by $\Phi_{S_1}$ then the feature maps after convolutions are shifted by $\Phi_{S_2}$ in addition to the original bias $b$. 
Detailed steps of \emph{SS} are given in Algorithm~\ref{alg_Meta} in Section~\ref{sec_alg}.

Figure~\ref{main_framework_SS_FT}(a) shows a typical parameter-level \emph{Fine-Tuning} (\emph{FT}) operation, which is in the meta optimization phase of our related work MAML~\cite{FinnAL17}.
It is obvious that \emph{FT} updates the complete values of $W$ and $b$, and has a large number of parameters, and our \emph{SS} reduces this number to below $\tfrac{2}{9}$ in the example of the figure. 

In summary, \emph{SS} can benefit MTL in three aspects.
1) It starts from a strong initialization  based on a large-scale trained DNN, yielding fast convergence for MTL.
2) It does not change DNN weights, thereby avoiding the problem of ``catastrophic forgetting''~\cite{LopezPazNIPS17, McCloskey1989} when learning specific tasks in MTL.
3) It is light-weight, reducing the chance of overfitting of MTL in few-shot scenarios.

\subsection{Hard task (HT) meta-batch}
\label{sec_HT}

In this section, we introduce a method to schedule hard tasks in meta-training batches. 
The conventional meta-batch is composed of randomly sampled tasks, where the randomness implies random difficulties~\cite{FinnAL17}.
In our meta-training pipeline, we intentionally pick up failure cases in each task and re-compose their data to be harder tasks for adverse re-training. 
We aim to force our meta-learner to ``grow up through hardness''. 

\myparagraph{Pipeline.}
Each task $\mathcal{T}$ has two splits, $\mathcal{T}^{(tr)}$ and $\mathcal{T}^{(te)}$, for base-learning and test, respectively.
As shown in Algorithm~\ref{alg_Meta} line 2-5, base-learner is optimized by the loss of $\mathcal{T}^{(tr)}$ (in multiple epochs). \emph{SS} parameters are then optimized by the loss of $\mathcal{T}^{(te)}$ once.
We can also get the recognition accuracy of $\mathcal{T}^{(te)}$ for $M$ classes. Then, we choose the lowest accuracy $Acc_m$ to determine the most difficult class-$m$ (also called failure class) in the current task.

After obtaining all failure classes (indexed by $\{m\}$) from $k$ tasks in current meta-batch $\{\mathcal{T}_{1\sim k}\}$, we re-sample tasks from their data. 
Specifically, we assume $p(\mathcal{T}|\{m\})$ is the task distribution, we sample a ``harder'' task $\mathcal{T}^{hard} \in p(\mathcal{T}|\{m\})$.
Two important details are given below.

\myparagraph{Choosing hard class-$m$.} We choose the failure class-$m$ from each task by ranking the class-level accuracies instead of fixing a threshold.
In a dynamic online setting as ours, it is more sensible to choose the hardest cases based on ranking rather than fixing a threshold ahead of time. 

\myparagraph{Two methods of hard tasking using $\{m\}$.}
Chosen $\{m\}$, we can re-sample tasks $\mathcal{T}^{hard}$ by (1) directly using the samples of class-$m$ in the current task $\mathcal{T}$, or (2) indirectly using the label of class-$m$ to sample new samples of that class.
In fact, setting (2) considers to include more data variance of class-$m$ and it works better than setting (1) in general. 

\subsection{Algorithm}
\label{sec_alg}

Algorithm~\ref{alg_overall} summarizes the training process of two main stages: large-scale DNN training (line 1-5) and meta-transfer learning (line 6-22). HT meta-batch re-sampling and continuous training phases are shown in lines 16-20, for which the failure classes are returned by Algorithm~\ref{alg_Meta}, see line 14.
Algorithm~\ref{alg_Meta} presents the learning process on a single task that includes episode training (lines 2-5) and episode test, i.e. meta-level update (lines 6). In lines 7-11, the recognition rates of all test classes are computed and returned to Algorithm~\ref{alg_overall} (line 14) for hard task sampling.


\begin{algorithm}
\caption{Meta-transfer learning (MTL)}
\label{alg_overall}
\SetAlgoLined
\SetKwInput{KwData}{Input}
\SetKwInput{KwResult}{Output}
 \KwData{Task distribution $p(\mathcal{T})$ and corresponding dataset $\mathcal{D}$, learning rates $\alpha$, $ \beta$ and $\gamma$}
 \KwResult{Feature extractor $\Theta$, base learner $\theta$, \emph{SS} parameters $\Phi_{S_{\{1,2\}}}$}
 Randomly initialize $\Theta$ and $\theta$\;
 \For{samples in $\mathcal{D}$}{
 Evaluate $\mathcal{L}_{\mathcal{D}}([\Theta; \theta])$ by Eq.~\ref{eq_large_scale_loss}\;
 Optimize $\Theta$ and $\theta$ by Eq.~\ref{eq_large_scale_update}\;
 }
  Initialize $\Phi_{S_1}$ by ones, initialize $\Phi_{S_2}$ by zeros\;
  Reset and re-initialize $\theta$ for few-shot tasks\;
 \For{meta-batches}{
 Randomly sample tasks $\{\mathcal{T}\}$ from $p(\mathcal{T})$\;
 \While{not done}{
 Sample task $\mathcal{T}_i \in \{\mathcal{T}$\}\;
 Optimize $\Phi_{S_{\{1,2\}}}$ and $\theta$ with $\mathcal{T}_i$ by \textbf{Algorithm}~\ref{alg_Meta}\;
 Get the returned class-$m$ then add it to $\{m\}$\;
 }
 Sample hard tasks $\{\mathcal{T}^{hard}\}$ from $\subseteq p(\mathcal{T}|\{m\})$\;
  \While{not done}{
 Sample task $\mathcal{T}^{hard}_j \in \{\mathcal{T}^{hard}$\} \;
 Optimize $\Phi_{S_{\{1,2\}}}$ and $\theta$ with $\mathcal{T}^{hard}_j$ by \textbf{Algorithm}~\ref{alg_Meta} \;
 }
 Empty $\{m\}$.
 }
\end{algorithm}



\begin{algorithm}
\caption{Detail learning steps within a task $\mathcal{T}$}
\label{alg_Meta}
\SetAlgoLined
\SetKwInput{KwData}{Input}
\SetKwInput{KwResult}{Output}
 \KwData{$\mathcal{T}$, learning rates $ \beta$ and $\gamma$, feature extractor $\Theta$, base learner $\theta$, \emph{SS} parameters $\Phi_{S_{\{1,2\}}}$}
 \KwResult{Updated $\theta$ and $\Phi_{S_{\{1,2\}}}$, 
 the worst classified class-$m$ in $\mathcal{T}$}
 Sample $\mathcal{T}^{(tr)}$ and $\mathcal{T}^{(te)}$ from $\mathcal{T}$ \;
 \For{samples in $\mathcal{T}^{(tr)}$}{
 Evaluate $\mathcal{L}_{\mathcal{T}^{(tr)}}$\;
 Optimize $\theta'$ by Eq.~\ref{eq_base_classifier}\;
 }
 Optimize $\Phi_{S_{\{1,2\}}}$ and $\theta$ by Eq.~\ref{eq_ss_update} and Eq.~\ref{eq_meta_classifier}\;
 \While{not done}{
 Sample class-$k$ in $\mathcal{T}^{(te)}$\;
 Compute $Acc_k$ for $\mathcal{T}^{(te)}$\;
 }
 Return class-$m$ with the lowest accuracy $Acc_m$.
\end{algorithm}
\vspace{-6mm}

\section{Experiments}
\label{sec_exp}

We evaluate the proposed \textbf{MTL} and \textbf{HT meta-batch} in terms of few-shot recognition accuracy and model convergence speed.
Below we describe the datasets and detailed settings, followed by an ablation study and a comparison to state-of-the-art methods.

\subsection{Datasets and implementation details}
\label{sec_dataset}

We conduct few-shot learning experiments on two benchmarks, miniImageNet~\cite{VinyalsBLKW16} and Fewshot-CIFAR100 (FC100)~\cite{OreshkinNIPS18}. 
miniImageNet is widely used in related works~\cite{FinnAL17, RaviICLR2017, GrantICLR2018, FranceschiICML18, MunkhdalaiICML18}. FC100 is newly proposed in~\cite{OreshkinNIPS18} and is more challenging in terms of lower image resolution and stricter training-test splits than miniImageNet.

\myparagraph{miniImageNet} was proposed by Vinyals \emph{et al}.~\cite{VinyalsBLKW16} for few-shot learning evaluation. 
Its complexity is high due to the use of ImageNet images, but requires less resource and infrastructure than running on the full ImageNet dataset~\cite{Russakovsky2015}. 
In total, there are $100$ classes with $600$ samples of $84 \times 84$ color images per class.
These $100$ classes are divided into $64$, $16$, and $20$ classes respectively for sampling tasks for meta-training, meta-validation and meta-test, following related works~\cite{FinnAL17, RaviICLR2017, GrantICLR2018, FranceschiICML18, MunkhdalaiICML18}.

\myparagraph{Fewshot-CIFAR100 (FC100)}
is based on the popular object classification dataset CIFAR100 \cite{CIFAR100}. The splits were proposed by~\cite{OreshkinNIPS18} (Please check details in the supplementary).
It offers a more challenging scenario with lower image resolution and more challenging meta-training/test splits that are separated according to object super-classes.
It contains $100$ object classes and each class has $600$ samples of $32 \times 32$ color images.
The $100$ classes belong to $20$ super-classes. Meta-training data are from $60$ classes belonging to $12$ super-classes. Meta-validation and meta-test sets contain $20$ classes belonging to $4$ super-classes, respectively. 
These splits accord to super-classes, thus minimize the information overlap between training and val/test tasks. 

The following settings are used on both datasets.
We train a large-scale DNN with all training datapoints (Section~\ref{sec_large_scale_pretrain}) and stop this training after $10k$ iterations.
We use the same task sampling method as related works~\cite{FinnAL17, RaviICLR2017}. Specifically, 1) we consider the 5-class classification and 2) we sample 5-class, 1-shot (5-shot or 10-shot) episodes to contain 1 (5 or 10) samples for train episode, and $15$ (uniform) samples for episode test.
Note that in the state-of-the-art work~\cite{OreshkinNIPS18}, $32$ and $64$ samples are respectively used in 5-shot and 10-shot settings for episode test.
In total, we sample $8k$ tasks for meta-training (same for w/ and w/o HT meta-batch), and respectively sample $600$ random tasks for meta-validation and meta-test. 
Please check the supplementary document (or \href{https://github.com/y2l/meta-transfer-learning-tensorflow}{GitHub repository}) for other implementation details, e.g. learning rate and dropout rate.

\myparagraph{Network architecture.} 
We present the details for
the Feature Extractor $\Theta$, 
MTL meta-learner with \emph{Scaling} $\Phi_{S_1}$ and \emph{Shifting} $\Phi_{S_2}$, 
and MTL base-learner (classifier) $\theta$. 

\myparagraph{The architecture of $\Theta$} have two options, ResNet-12 and 4CONV, commonly used in related works~\cite{FinnAL17, VinyalsBLKW16, RaviICLR2017, MunkhdalaiICML18, MishraICLR2018, OreshkinNIPS18}.
\myparagraph{4CONV} consists of $4$ layers with $3\times 3$ convolutions and $32$ filters, followed by batch normalization (BN)~\cite{IoffeICML15}, a ReLU nonlinearity, and $2\times 2$ max-pooling.
\myparagraph{ResNet-12} is more popular in recent works~\cite{OreshkinNIPS18, MishraICLR2018, FranceschiICML18, MunkhdalaiICML18}. It contains $4$ residual blocks and each block has $3$ CONV layers with $3\times 3$ kernels.
At the end of each residual block, a $2\times 2$ max-pooling layer is applied. The number of filters starts from $64$ and is doubled every next block. 
Following $4$ blocks, there is a mean-pooling layer to compress the output feature maps to a feature embedding.
\myparagraph{The difference} between using 4CONV and using ResNet-12 in our methods is that ResNet-12 MTL sees the large-scale data training, but 4CONV MTL is learned from scratch because of its poor performance for large-scale data training (see results in the supplementary).
Therefore, we emphasize the experiments of using ResNet-12 MTL for its superior performance.
\myparagraph{The architectures of $\Phi_{S_1}$ and $\Phi_{S_2}$} are generated according to the architecture of $\Theta$, as introduced in Section~\ref{sec_meta_transfer}. That is when using ResNet-12 in MTL, $\Phi_{S_1}$ and $\Phi_{S_2}$ also have 12 layers, respectively.
\myparagraph{The architecture of $\theta$} is an FC layer. We empirically find that a single FC layer is faster to train and more effective for classification than multiple layers.
(see comparisons in the supplementary).

\begin{figure*}
\includegraphics[height=1.09in]{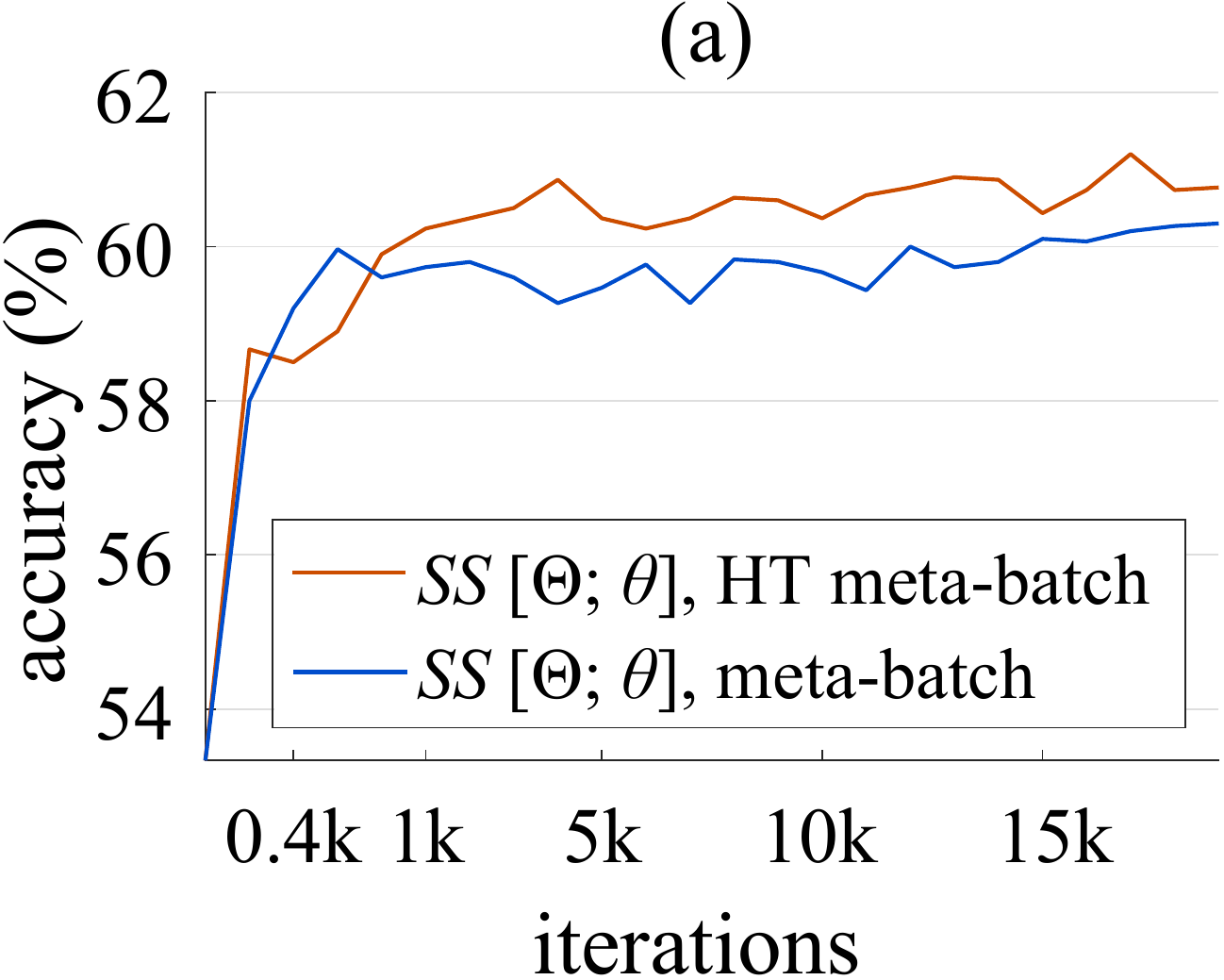}
\includegraphics[height=1.09in]{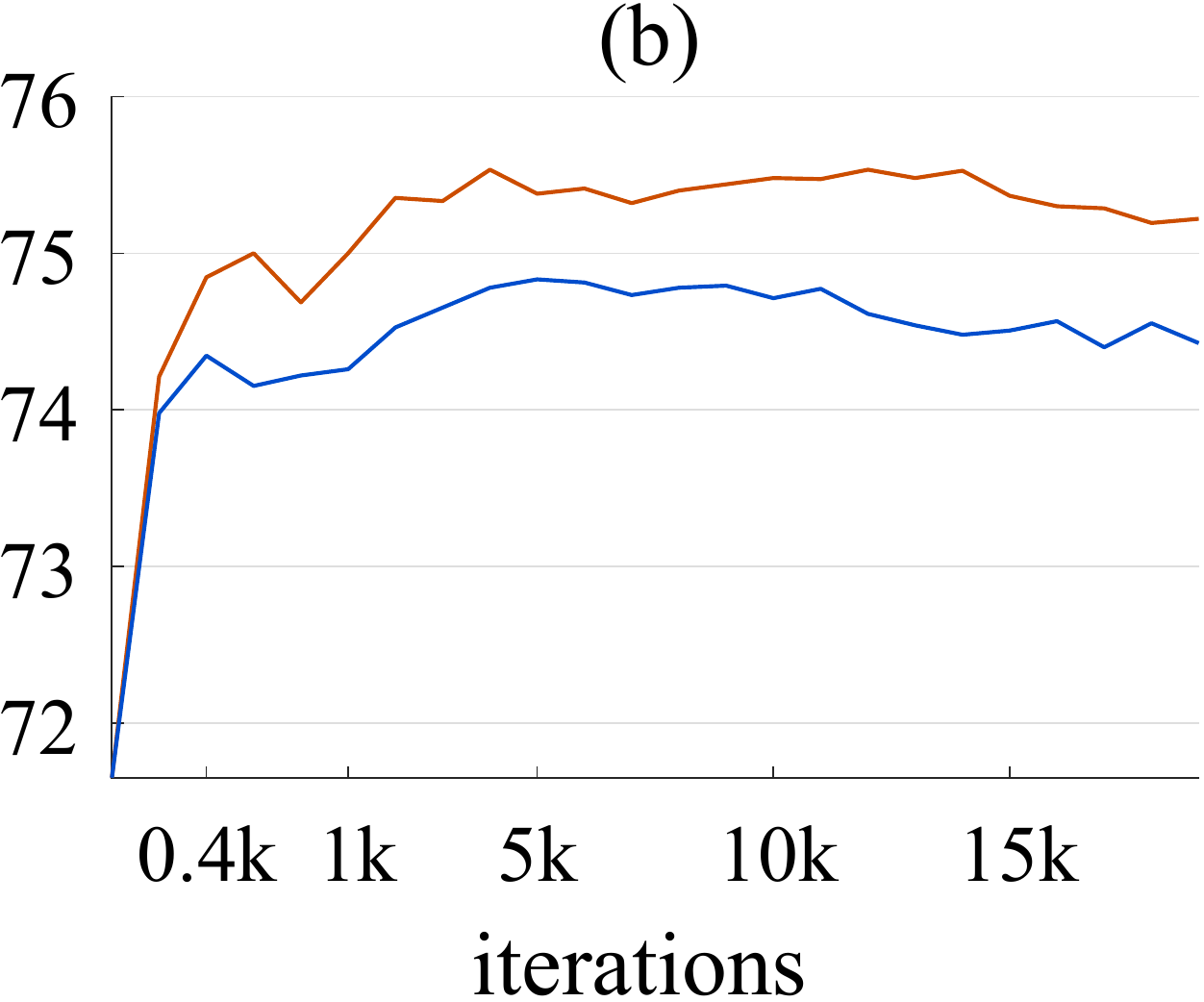}
\includegraphics[height=1.09in]{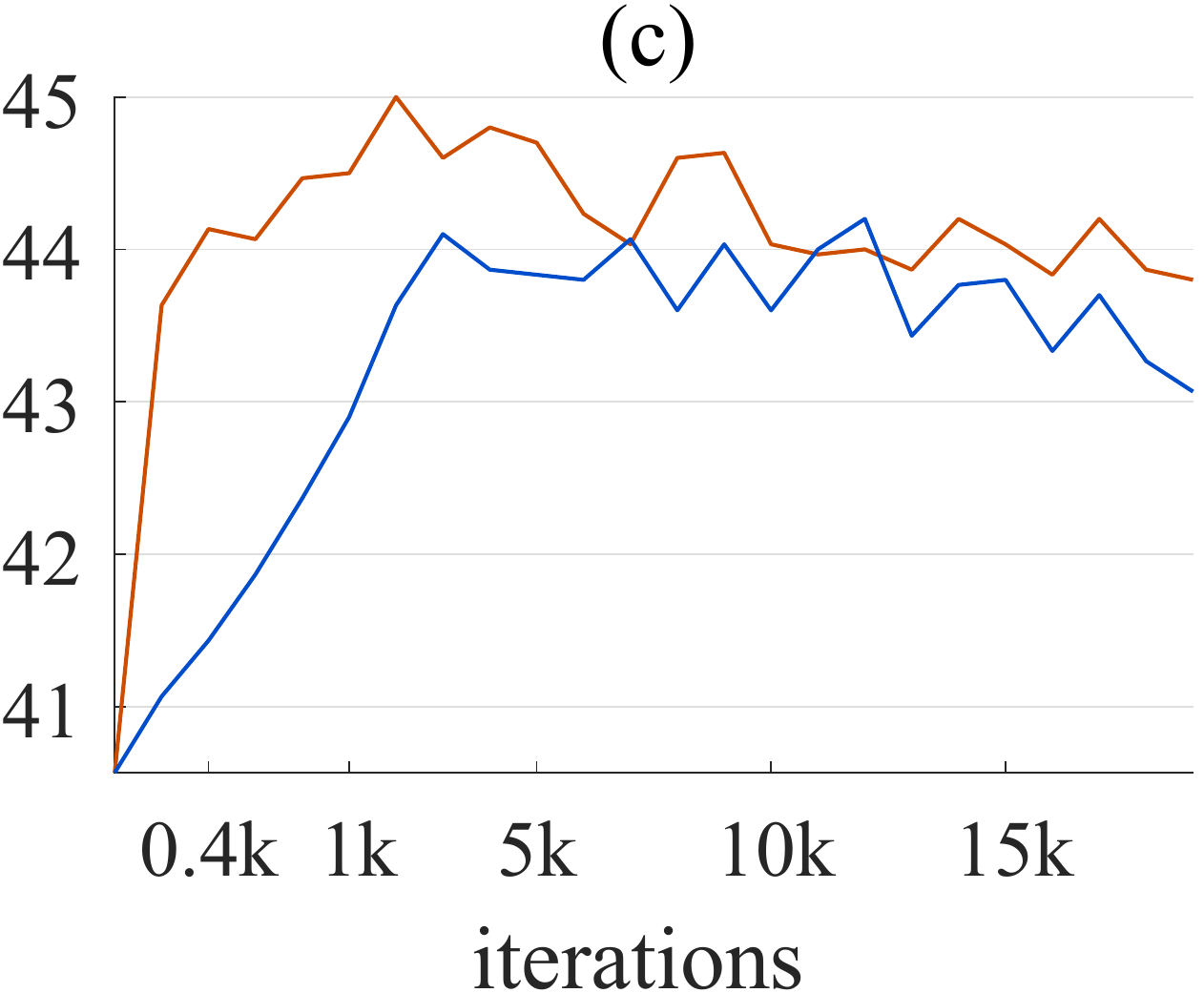}
\includegraphics[height=1.09in]{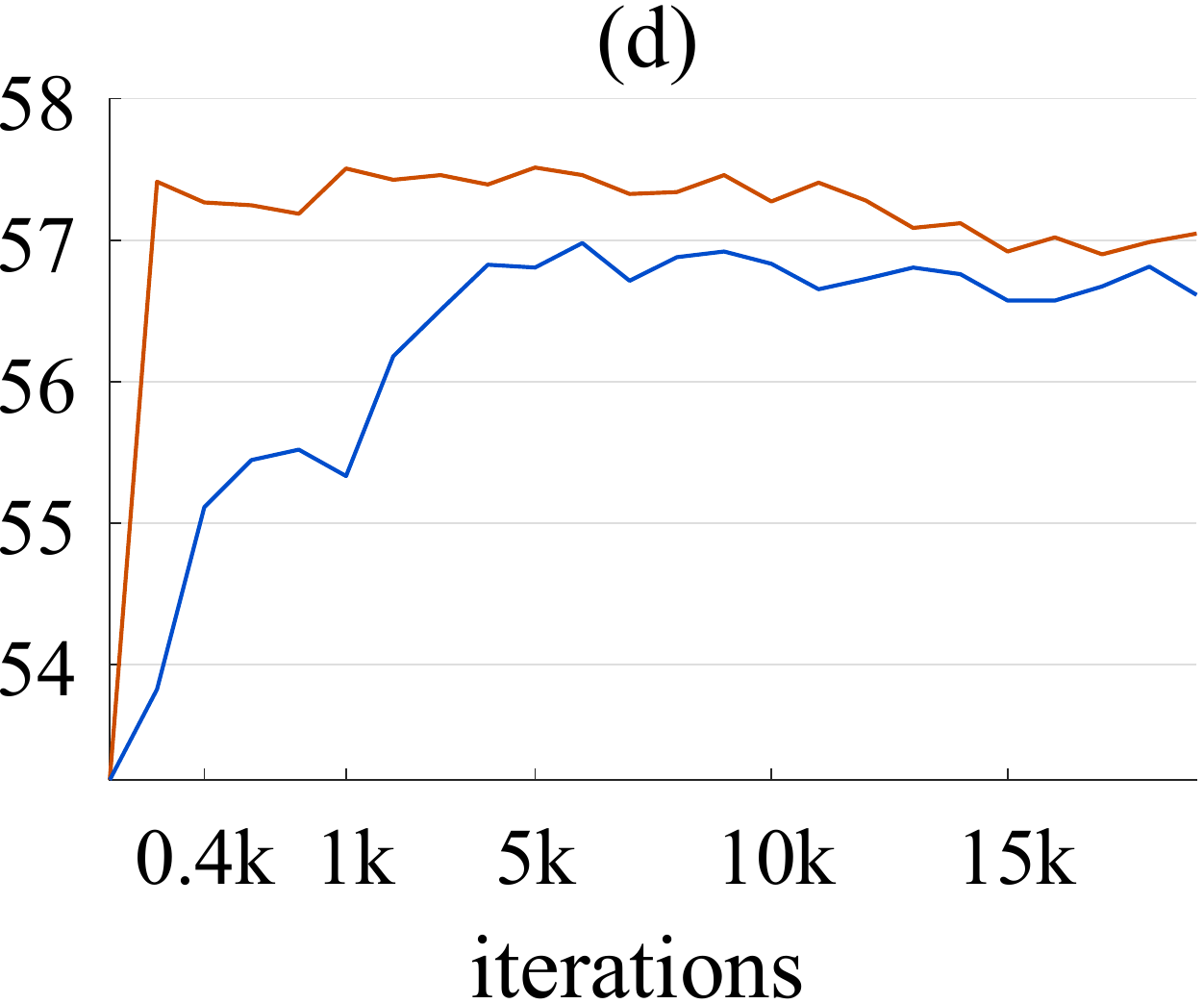}
\includegraphics[height=1.09in]{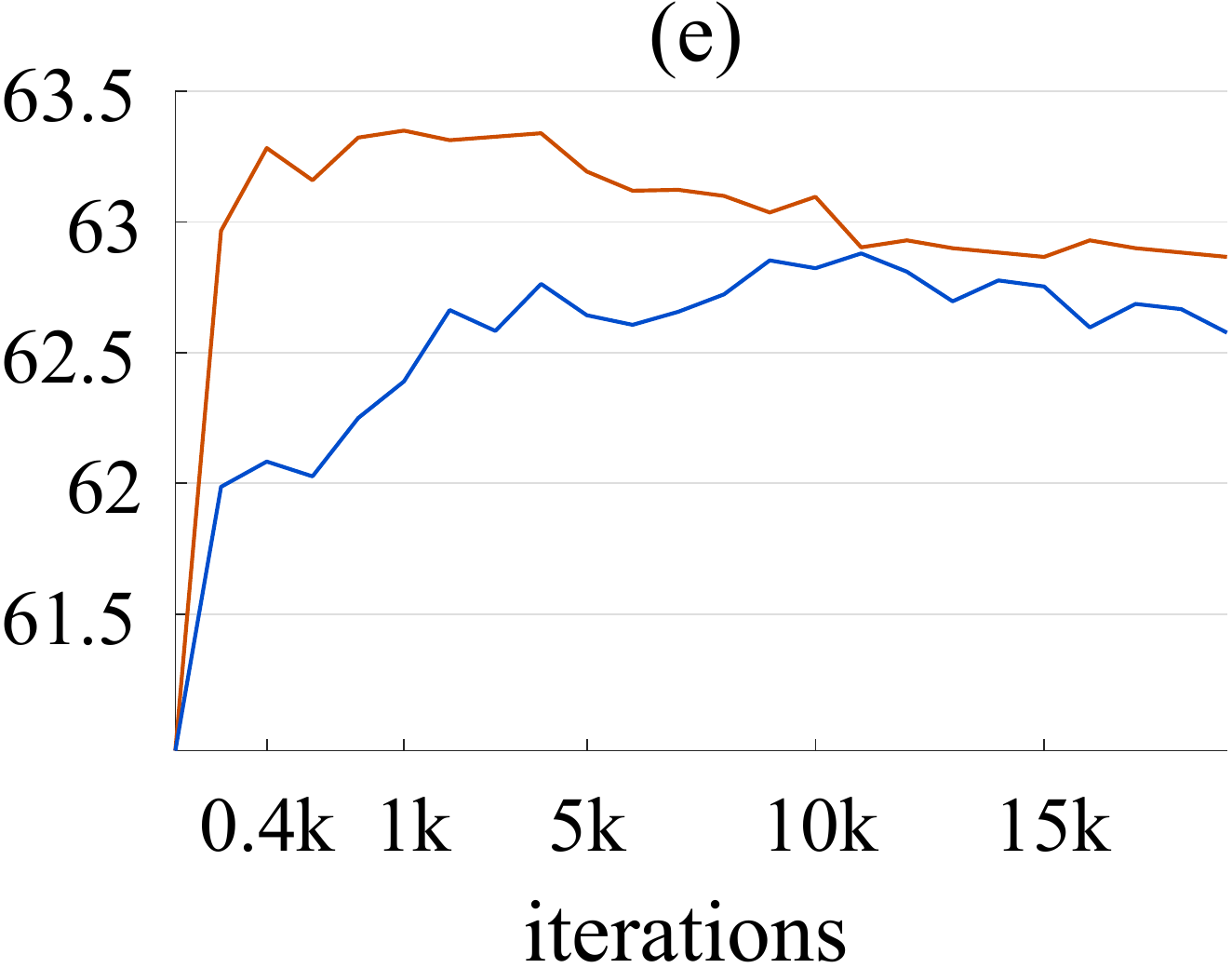}
\caption{(a)(b) show the results of 1-shot and 5-shot on miniImageNet; (c)(d)(e) show the results of 1-shot, 5-shot and 10-shot on FC100.}
\vspace{-0.3cm}
\label{fig_mini_fc100}
\end{figure*}

\subsection{Ablation study setting}
\label{sec_setting}

In order to show the effectiveness of our approach, we design some ablative settings: 
two baselines without meta-learning but more classic learning,
three baselines of \emph{Fine-Tuning} (\emph{FT}) on smaller number of parameters (Table~\ref{tab_ablation_study}), and
two MAML variants using our deeper pre-trained model and HT meta-batch (Table~\ref{table_mini} and Table~\ref{table_fc100}).
Note that the alternative meta-learning operation to \emph{SS} is the \emph{FT} used in MAML.
Some bullet names are explained as follows.

\myparagraph{\emph{update} $[\Theta; \theta]$ (or $\theta$).} 
There is no meta-training phase. During test phase, each task has its whole model $[\Theta; \theta]$ (or the classifier $\theta$) updated on $\mathcal{T}^{(tr)}$, and then tested on $\mathcal{T}^{(te)}$.

\myparagraph{\emph{FT} $[\Theta 4; \theta]$ (or $\theta$).}
These are straight-forward ways to define a smaller set of meta-learner parameters than MAML.
We can freeze low-level pre-trained layers and meta-learn the classifier layer $\theta$ with (or without) high-level CONV layer $\Theta 4$ that is the 4th residual block of ResNet-12.

\subsection{Results and analysis}
\label{sec_result}

\begin{table}[b]
\centering
\vspace{-0.4cm}
\small
\begin{tabular*}{8cm}
{lcccccc}
\toprule 
& \multicolumn{2}{c}{miniImageNet} & & \multicolumn{3}{c}{FC100} \\
\cmidrule{2-3}\cmidrule{5-7}
& 1 (shot) & 5 & & 1 & 5 & 10 \\
\midrule[1pt]
\emph{update} $[\Theta; \theta]$ & 45.3 & 64.6  & & 38.4 & 52.6 & 58.6 \\
\emph{update} $\theta$ & 50.0 & 66.7  & & 39.3 & 51.8 & 61.0 \\
\midrule[1pt]	
\emph{FT} $\theta$  & 55.9 & 71.4  & & 41.6 & 54.9 & 61.1 \\
\emph{FT} $[\Theta4; \theta]$ & 57.2 & 71.6 & & 40.9 & 54.3 & 61.3 \\
\emph{FT} $[\Theta; \theta]$  & 58.3 & 71.6  & & 41.6 & 54.4 & 61.2 \\
\midrule[1pt]
\emph{SS} $[\Theta4; \theta]$ & 59.2 & 73.1  & &42.4 &55.1 &61.6 \\
\emph{SS $[\Theta; \theta]$}\textbf{(Ours)} & 60.2 & 74.3 & & 43.6 & 55.4 & 62.4  \\
\bottomrule[1pt]
\end{tabular*}
\vspace{0.1cm}
\caption{Classification accuracy (\%) using ablative models, on two datasets. ``meta-batch'' and ``ResNet-12(pre)'' are used.}
\label{tab_ablation_study}
\end{table}

\begin{table*}
  \small
  \centering
  \begin{tabular}{l l lcc}
    \toprule
     \multicolumn{2}{c}{Few-shot learning method} & Feature extractor & 1-shot & 5-shot \\
    \midrule
    \multirow{2}{*}{\emph{Data augmentation}}
    & Adv. ResNet, \cite{Mehrotra2017} & WRN-40 (pre) & 55.2 & 69.6 \\
    & Delta-encoder, \cite{SchwartzNIPS18} & VGG-16 (pre) & 58.7 & 73.6 \\
    \midrule  
    \multirow{3}{*}{\emph{Metric learning}}
    &Matching Nets, \cite{VinyalsBLKW16} & 4 CONV & 43.44 $\pm$ $0.77$ & 55.31 $\pm$  $0.73$ \\
    &ProtoNets, \cite{SnellSZ17} & 4 CONV & 49.42 $\pm$ $0.78$ & 68.20 $\pm$ $0.66$\\
    &CompareNets, \cite{SungCVPR2018} & 4 CONV & 50.44 $\pm$ $0.82$ & 65.32 $\pm$ $0.70$\\
         \midrule
    \multirow{3}{*}{\emph{Memory network}} 
    & Meta Networks, \cite{MunkhdalaiICML2017} & 5 CONV  & 49.21 $\pm$ $0.96$ & -- \\
    & SNAIL, \cite{MishraICLR2018} & ResNet-12 (pre)${}^{\diamond}$  & 55.71 $\pm$ $0.99$  & 68.88 $\pm$ $0.92$\\
    & TADAM, \cite{OreshkinNIPS18} & ResNet-12 (pre)${}^{\dag}$  & 58.5 $\pm$ $0.3$  & \textbf{76.7} $\mathbf{\pm}$ $\mathbf{0.3}$\\
    \midrule
    \multirow{6}{*}{\emph{Gradient descent}}
    & MAML, \cite{FinnAL17} & 4 CONV & 48.70 $\pm$ $1.75$ & 63.11 $\pm$ $0.92$ \\
    & Meta-LSTM, \cite{RaviICLR2017} & 4 CONV & 43.56 $\pm$ $0.84$ & 60.60 $\pm$ $0.71$ \\
    & Hierarchical Bayes, \cite{GrantICLR2018} & 4 CONV  & 49.40 $\pm$ $1.83$ & -- \\
    & Bilevel Programming, \cite{FranceschiICML18} & ResNet-12${}^{\diamond}$   & 50.54 $\pm$ $0.85$  & 64.53 $\pm$ $0.68$\\
    & MetaGAN, \cite{ZhangNIPS2018MetaGAN} & ResNet-12 & 52.71 $\pm$ $0.64$  & 68.63 $\pm$ $0.67$ \\
    & adaResNet, \cite{MunkhdalaiICML18} & ResNet-12${}^{\ddag}$   & 56.88 $\pm$ $0.62$ & 71.94 $\pm$ $0.57$ \\
    \midrule
    MAML, HT  & \emph{FT} $[\Theta; \theta]$, HT meta-batch & 4 CONV & 49.1 $\pm$ $1.9$ & 64.1 $\pm$ $0.9$ \\
    MAML deep, HT  & \emph{FT} $[\Theta; \theta]$, HT meta-batch & ResNet-12 (pre) & 59.1 $\pm$ $1.9$ & 73.1 $\pm$ $0.9$ \\
    \midrule
    \multirow{2}{*}{\textbf{MTL (Ours)}}
     & \emph{SS} $[\Theta; \theta]$, meta-batch & ResNet-12 (pre) & 60.2 $\pm$ $1.8$ & 74.3 $\pm$ $0.9$\\
     & \emph{SS} $[\Theta; \theta]$, HT meta-batch & ResNet-12 (pre) & \textbf{61.2} $\mathbf{\pm}$ $\mathbf{1.8}$ & \textbf{75.5} $\mathbf{\pm}$ $\mathbf{0.8}$\\
  \bottomrule
    \multicolumn{5}{l}{${}^{\diamond}$Additional 2 convolutional layers { } ${}^{\ddag}$Additional 1 convolutional layer { } ${}^{\dag}$Additional 72 fully connected layers}\\
\end{tabular}
  \vspace{0.2cm}
  \caption{The 5-way, 1-shot and 5-shot classification accuracy ($\%$) on miniImageNet dataset. ``pre'' means pre-trained for a single classification task using all training datapoints.}
    \label{table_mini}
\end{table*}

\begin{table*}[t]
  \small
  \centering
  \begin{tabular}{l l lccc}
    \toprule
     \multicolumn{2}{c}{Few-shot learning method} & Feature extractor & 1-shot & 5-shot & 10-shot \\
    \midrule    
    \multirow{1}{*}{\emph{Gradient descent}} 
    & MAML, ~\cite{FinnAL17}${}^{\ddag}$ & 4 CONV  &  38.1 $\pm$ $1.7$ & 50.4 $\pm$ $1.0$  & 56.2 $\pm$ $0.8$\\
    \midrule
    \multirow{1}{*}{\emph{Memory network}}
    & TADAM, \cite{OreshkinNIPS18} & ResNet-12 (pre)${}^{\dag}$   & 40.1 $\pm$ $0.4$  & 56.1 $\pm$ $0.4$ &  61.6 $\pm$ $0.5$\\
    \midrule
   MAML, HT  & \emph{FT} $[\Theta; \theta]$, HT meta-batch & 4 CONV & 39.9 $\pm$ $1.8$ & 51.7 $\pm$ $0.9$ &  57.2 $\pm$ $0.8$ \\
    MAML deep, HT  &  \emph{FT} $[\Theta; \theta]$, HT meta-batch & ResNet-12 (pre) & 41.8 $\pm$ $1.9$ & 55.1 $\pm$ $0.9$ & 61.9 $\pm$ $0.8$ \\
     \midrule
    \multirow{2}{*}{\textbf{MTL (Ours)}}
    & \emph{SS} $[\Theta; \theta]$, meta-batch & ResNet-12 (pre) & 43.6 $\pm$ $1.8$ & 55.4 $\pm$ $0.9$ & 62.4 $\pm$ $0.8$ \\
    & \emph{SS} $[\Theta; \theta]$, HT meta-batch & ResNet-12 (pre) & \textbf{45.1} $\mathbf{\pm}$ $\mathbf{1.8}$ &  \textbf{57.6} $\mathbf{\pm}$ $\mathbf{0.9}$ & \textbf{63.4} $\mathbf{\pm}$ $\mathbf{0.8}$ \\
  \bottomrule
  \multicolumn{6}{l}{${}^{\dag}$Additional 72 fully connected layers  { } ${}^{\ddag}$Our implementation using the public code of MAML.}\\
\end{tabular}
  \vspace{0.1cm}
  \caption{The 5-way with 1-shot, 5-shot and 10-shot classification accuracy (\%) on Fewshot-CIFAR100 (FC100) dataset. ``pre'' means pre-trained for a single classification task using all training datapoints.}
    \label{table_fc100}
    \vspace{-0.3cm}
\end{table*}

Table~\ref{tab_ablation_study}, Table~\ref{table_mini} and Table~\ref{table_fc100} present the overall results on miniImageNet and FC100 datasets. 
Extensive comparisons are done with ablative methods and the state-of-the-arts. 
Note that tables present the highest accuracies for which the iterations were chosen
by validation. For the miniImageNet, iterations for 1-shot and 5-shot are at $17k$ and $14k$, respectively. For the FC100, iterations are all at $1k$.
Figure~\ref{fig_mini_fc100} shows the performance gap between \emph{with} and \emph{without} HT meta-batch in terms of accuracy and converging speed.

\myparagraph{Result overview on miniImageNet.} 
In Table~\ref{table_mini}, we can see that the proposed MTL with \emph{SS} $[\Theta; \theta]$, HT meta-batch and ResNet-12(pre) achieves the best few-shot classification performance with $61.2\%$ for (5-class, 1-shot). Besides, it tackles the (5-class, 5-shot) tasks with an accuracy of $75.5\%$ that is comparable to the state-of-the-art results, i.e. $76.7\%$, reported by TADAM~\cite{OreshkinNIPS18} whose model used $72$ additional FC layers in the ResNet-12 arch.
In terms of the network arch, it is obvious that
models using ResNet-12 (pre) outperforms those using 4CONV by large margins, e.g. 4CONV models have the best 1-shot result with $50.44\%$~\cite{SungCVPR2018} which is $10.8\%$ lower than our best.

\myparagraph{Result overview on FC100.}
In Table~\ref{table_fc100}, we give the results of TADAM using their reported numbers in the paper~\cite{OreshkinNIPS18}. 
We used the public code of MAML~\cite{FinnAL17} to get its results for this new dataset. 
Comparing these methods, we can see that MTL consistently outperforms MAML by large margins, i.e. around $7\%$ in all tasks; and surpasses TADAM by a relatively larger number of $5\%$ for 1-shot, and with $1.5\%$ and $1.8\%$ respectively for 5-shot and 10-shot tasks. 

\myparagraph{MTL \emph{vs}. No meta-learning.}
Table~\ref{tab_ablation_study} shows the results of \emph{No meta-learning} on the top block.
Compared to these, our approach achieves significantly better performance even \emph{without} HT meta-batch,
e.g. the largest margins are $10.2\%$ for 1-shot and $8.6\%$ for 5-shot on miniImageNet.
This validates the effectiveness of our meta-learning method for tackling few-shot learning problems.
Between two \emph{No meta-learning} methods, we can see that updating both feature extractor $\Theta$ and classifier $\theta$ is inferior to updating $\theta$ only, e.g. around $5\%$ reduction on miniImageNet 1-shot.
One reason is that in few-shot settings, there are too many parameters to optimize with little data.
This supports our motivation to learn only $\theta$ during base-learning.

\myparagraph{Performance effects of MTL components.} 
MTL with full components, \emph{SS} $[\Theta; \theta]$, HT meta-batch and ResNet-12(pre), achieves the best performances for all few-shot settings on both datasets, see Table~\ref{table_mini} and Table~\ref{table_fc100}. 
We can conclude that our large-scale network training on deep CNN significantly boost the few-shot learning performance. This is an important gain brought by the transfer learning idea in our MTL approach. 
It is interesting to note that this gain on FC100 is not as large as for miniImageNet: only $1.7\%$, $1.0\%$ and $4.0\%$. 
The possible reason is that FC100 tasks for meta-train and meta-test are clearly split according to super-classes. The data domain gap is larger than that for miniImageNet, which makes transfer more difficult.

HT meta-batch and ResNet-12(pre) in our approach can be generalized to other meta-learning models. 
MAML 4CONV with HT meta-batch gains averagely $1\%$ on two datasets. When changing 4CONV by deep ResNet-12 (pre)
it achieves significant improvements, e.g. $10\%$ and $9\%$ on miniImageNet.
Compared to MAML variants, our MTL results are consistently higher, e.g. $2.5\% \sim 3.3\%$ on FC100.
People may argue that MAML fine-tuning(\emph{FT}) all network parameters is likely to overfit to few-shot data. In the middle block of Table~\ref{tab_ablation_study},
we show the ablation study of freezing low-level pre-trained layers and meta-learn only the high-level layers (e.g. the $4$-th residual block of ResNet-12) by the \emph{FT} operations of MAML.
These all yield inferior performances than using our \emph{SS}. 
An additional observation is that \emph{SS}* performs consistently better than \emph{FT}*.

\myparagraph{Speed of convergence of MTL.}
MAML~\cite{FinnAL17} used $240k$ tasks to achieve the best performance on miniImageNet.
Impressively, our MTL methods used only $8k$ tasks, see Figure~\ref{fig_mini_fc100}(a)(b) (note that each iteration contains 2 tasks). 
This advantage is more obvious for FC100 on which MTL methods need at most $2k$ tasks, Figure~\ref{fig_mini_fc100}(c)(d)(e).
We attest this to two reasons. First, MTL starts from the pre-trained ResNet-12. 
And second, \emph{SS} (in MTL) needs to learn only $< \tfrac{2}{9}$ parameters of the number of \emph{FT} (in MAML) when using ResNet-12. 

\myparagraph{Speed of convergence of HT meta-batch.} 
Figure~\ref{fig_mini_fc100} shows 1) MTL with HT meta-batch consistently achieves higher performances than MTL with the conventional meta-batch~\cite{FinnAL17}, in terms of the recognition accuracy in all settings; 
and 2) it is impressive that MTL with HT meta-batch achieves top performances early, after \eg about $2k$ iterations for 1-shot, $1k$ for 5-shot and $1k$ for 10-shot, on the more challenging dataset -- FC100.

\section{Conclusions}

In this paper, we show that our novel MTL trained with HT meta-batch learning curriculum achieves the top performance for tackling few-shot learning problems.
The key operations of MTL on pre-trained DNN neurons proved highly efficient for adapting learning experience to the unseen task.
The superiority was particularly achieved in the extreme 1-shot cases on two challenging benchmarks -- miniImageNet and FC100. 
In terms of learning scheme, HT meta-batch showed consistently good performance for all baselines and ablative models.
On the more challenging FC100 benchmark, it showed to be particularly helpful for boosting convergence speed.
This design is independent from any specific model and could be generalized well whenever the hardness of task is easy to evaluate in online iterations.

\section*{Acknowledgments}
This research is part of NExT research which is supported by the National Research Foundation, Prime Minister's Office, Singapore under its IRC@SG Funding Initiative.
It is also partially supported by German Research Foundation (DFG CRC 1223), and National Natural Science Foundation of China (61772359).

{\small
\bibliographystyle{ieee}
\bibliography{egbib}
}

\clearpage

\beginsupp
\setcounter{section}{0}
\renewcommand\thesection{\Alph{section}}
\noindent
{\Large {\textbf{Supplementary materials}}}
\\

These materials include the details of network architecture (\S\ref{suppsec:arch}), implementation (\S\ref{suppsec:implementation}), FC100 dataset splits (\S\ref{suppsec:fc100splits}), standard variance analysis (\S\ref{suppsec:ci95}), additional ablation results (\S\ref{suppsec:addexp}), and some interpretation of our meta-learned model (\S\ref{sec_visual}).
In addition, our open-source code is on GitHub\footnote{\href{https://github.com/y2l/meta-transfer-learning-tensorflow}{https://github.com/y2l/meta-transfer-learning-tensorflow}}.


\section{Network architectures}
\label{suppsec:arch}
\begin{figure*}
\includegraphics[width=7in]{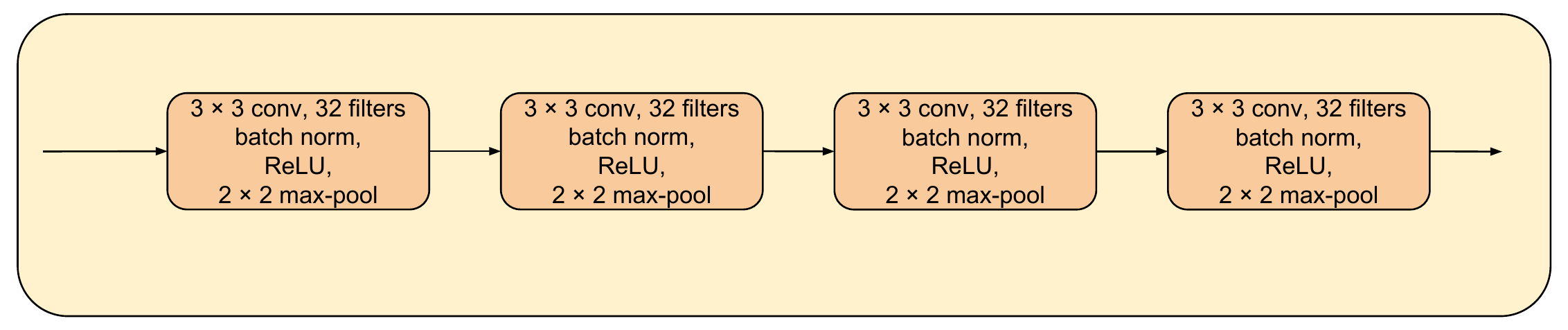}
\caption{Network architecture of 4CONV}
\label{figure_netarch_4CONV}
\end{figure*}

In Figure~\ref{figure_netarch_4CONV}, we present the 4CONV architecture for feature extractor $\Theta$, as illustrated in Section 5.1 ``Network architecture'' of the main paper.

\begin{figure*}
\includegraphics[width=7in]{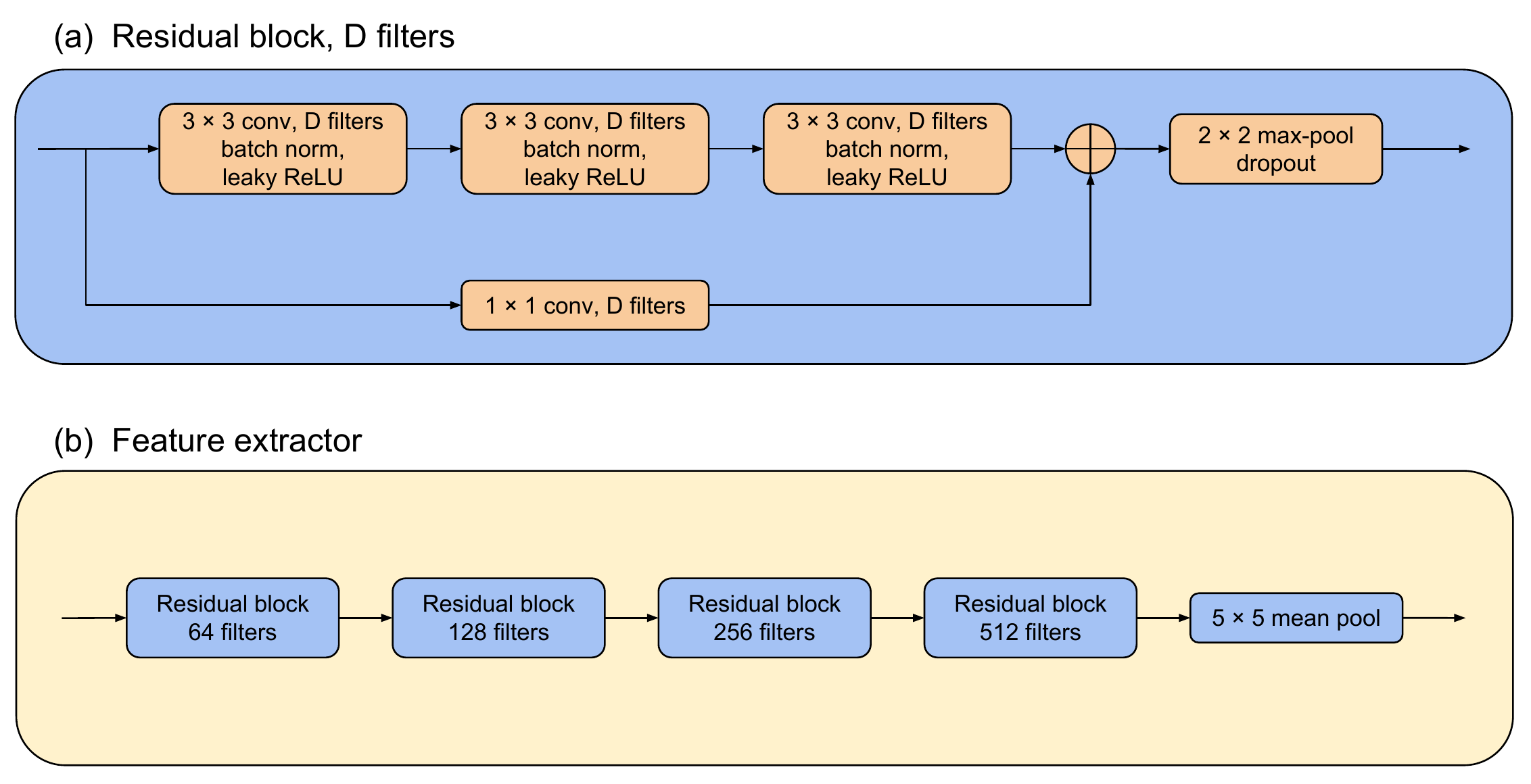}
\caption{Network architecture of ResNet-12}
\label{figure_netarch}
\end{figure*}

In Figure~\ref{figure_netarch}, we present the other architecture -- ResNet-12.
Figure~\ref{figure_netarch}(a) shows the details of a single residual block and Figure~\ref{figure_netarch}(b) shows the whole network consisting of four residual blocks and a mean-pooling layer.

The input of $\Theta$ is the $3$-channel RGB image, and the output is the $512$-dimensional feature vector. 
$a = 0.1$ is set for all leakyReLU activation functions in ResNet-12.

\section{Implementation details}
\label{suppsec:implementation}
For \textbf{the phase of DNN training on large-scale data}, the model is trained by Adam optimizer~\cite{kingma2014adam}. Its learning rate is initialized as $0.001$, and decays to its half every $5k$ iterations until it is lower that $0.0001$. 
We set the keep probability of the dropout as $0.9$ and batch-size as $64$.
The pre-training stops after $10k$ iterations.
Note that for the hyperparameter selection, we randomly choose $550$ samples each class as the training set, and the rest as validation. 
After the grid search of hyperparameters, we fix them and mix up all samples ($64$ classes, $600$ samples each class), in order to do the final pre-training.
Besides, these pre-training samples are augmented with horizontal flip.

For the \textbf{meta-train phase}, we sample 5-class, 1-shot (5-shot or 10-shot) episodes to contain $1$ ($5$ or $10$) sample(s) for episode training, and $15$ samples for episode test uniformly, following the setting of MAML~\cite{FinnAL17}. 
The base-learner $\theta$ is optimized by batch gradient descent with the learning rate of $0.01$. It gets updated with $20$ and $60$ epochs respectively for 1-shot and 5-shot tasks on the miniImageNet dataset, and $20$ epochs for all tasks on the FC100 dataset. 
The meta-learner, \ie, the parameters of the \emph{SS} operations, is optimized by Adam optimizer~\cite{kingma2014adam}.
Its learning rate is initialized as $0.001$, and decays to the half every $1k$ iterations until $0.0001$. 
The size of meta-batch is set to $2$ (tasks) due to the memory limit. 

Using our \textbf{HT meta-batch strategy}, hard tasks are sampled every time after running $10$ meta-batches, \ie, the failure classes used for sampling hard tasks are from $20$ tasks. The number of hard task is selected for different settings by validation: $10$ and $4$ hard tasks respectively for the 1-shot and 5-shot experiments on the miniImageNet dataset; and respectively $20$, $10$ and $4$ hard tasks for the 1-shot, 5-shot and 10-shot experiments on the FC100 dataset.

For the \textbf{meta-test phase}, we sample 5-class, 1-shot (5-shot or 10-shot) episodes and each episode contains $1$ ($5$ or $10$) sample(s) for both episode train and episode test. 
On each dataset, we sample $600$ meta-test tasks. All these settings are exactly the same as MAML~\cite{FinnAL17}.

\section{Super-class splits on FC100}
\label{suppsec:fc100splits}
In this section, we show the details of the FC100 splits according to the super-class labels, same with TADAM~\cite{OreshkinNIPS18}.

\myparagraph{Training split}
super-class indexes: 1, 2, 3, 4, 5, 6, 9, 10, 15, 17, 18, 19; and
corresponding labels: fish, flowers, food\_containers, fruit\_and\_vegetables, household\_electrical\_devices, household\_furniture, large\_man-made\_outdoor\_things, large\_natural\_outdoor\_scenes, reptiles, trees, vehicles\_1, vehicles\_2.

\myparagraph{Validation split}
super-class indexes: 8, 11, 13, 16; and
corresponding labels: large\_carnivores, large\_omnivores\_and\_herbivores, non-insect\_invertebrates, small\_mammals.

\myparagraph{Test split}
super-class indexes: 0, 7, 12, 14; and
corresponding labels: aquatic\_mammals, insects, medium\_mammals, people.

An episode (task) is independently sampled from a corresponding split, \eg a meta-train episode contains $5$ classes that can only be belonging to the $12$ super-classes in the training split. Therefore, there is no fine-grained information overlap between meta-train and meta-test tasks.

\section{Standard variance analysis}
\label{suppsec:ci95}
The final accuracy results reported in our main paper are the mean values and standard variances of the results of $600$ meta-test tasks.
The standard variance is affected by the number of episode test samples.
As introduced in \S\ref{suppsec:implementation}, we use the same setting as MAML~\cite{FinnAL17} which used a smaller number of samples for episode test ($1$ sample for 1-shot episode test and $5$ samples for 5-shot), making the result variance higher.
Other works that used more samples for episode test got lower variances, \eg, TADAM~\cite{OreshkinNIPS18} used $100$ samples and its variances are about $\tfrac{1}{6}$ and $\tfrac{1}{3}$ of MAML's respectively for miniImageNet 1-shot and 5-shot.

In order to have a fair comparison with TADAM in terms of this issue, we supplement
the experiments using $100$ episode test samples at the meta-test.
We get the new confidence intervals (using our method: MTL w/o HT meta-batch) as $0.71\%$ ($0.3\%$ for TADAM) and $0.54\%$ ($0.3\%$ for TADAM) respectively for 1-shot and 5-shot on the miniImageNet dataset, and $0.70\%$ ($0.4\%$ for TADAM), $0.63\%$ ($0.4\%$ for TADAM) and $0.58\%$ ($0.5\%$ for TADAM) respectively for 1-shot, 5-shot and 10-shot on the FC100 dataset.

\section{Additional ablation study}
\label{suppsec:addexp}
We supplement the results in Table~\ref{table}, for the comparisons mentioned in Section 5.1 of main paper. 
Red numbers on the bottom row are copied from the main paper (corresponding to the MTL setting: \emph{SS} $\Theta$, meta-batch) and shown here for the convenience of comparison.

To get the first row, we train 4CONV net by large-scale data (same to the pre-training of ResNet-12) and get inferior results, as we declared in the main paper.
Results on the second and third rows show the performance drop when changing the single FC layer $\theta$ to multiple layers, \eg $2$ FC layers and $3$ FC layers. 
Results on the fourth row show the performance drop when updating both $\Theta$ and $\theta$ for the base-learning. The reason is that $\Theta$ has too many parameters to update with too little data.

\begin{table*}[t]
  \small
  \centering
  \begin{tabular}{ ccclccccc}
    \toprule
      \multirow{2}{*}{Meta-learning} & \multirow{2}{*}{Base-learning} & \multirow{2}{*}{FC dim of $\theta$} &  \multirow{2}{*}{Feature extractor} & \multicolumn{2}{c}{miniImageNet} &  \multicolumn{3}{c}{FC100}\\
       & & & &  1-shot & 5-shot & 1-shot & 5-shot & 10-shot \\
    \midrule    

     $\Phi_{S_1}$, $\Phi_{S_2}$ & $\theta$ & 5 & 4 CONV (pre) & 45.6 $\pm$ $1.8$ & 61.2 $\pm$ $0.9$ & 38.0 $\pm$ $1.6$ & 46.4 $\pm$ $0.9$ & 56.5 $\pm$ $0.8$ \\
     \midrule
     $\Phi_{S_1}$, $\Phi_{S_2}$ & $\theta$ (2-layer) & 512, 5 & ResNet-12 (pre) & 59.1 $\pm$ $1.9$ & 70.7 $\pm$ $0.9$ & 40.3 $\pm$ $1.9$ & 53.3 $\pm$ $0.9$ & 54.1 $\pm$ $0.8$ \\
     $\Phi_{S_1}$, $\Phi_{S_2}$ & $\theta$ (3-layer) & 1024, 512, 5 & ResNet-12 (pre) & 56.2 $\pm$ $1.8$ & 68.7 $\pm$ $0.9$ & 40.0 $\pm$ $1.8$  & 52.3 $\pm$ $1.0$  & 53.8 $\pm$ $0.8$  \\
     \midrule
     $\Phi_{S_1}$, $\Phi_{S_2}$ & $\Theta$, $\theta$ & 5 & ResNet-12 (pre) & 59.6 $\pm$ $1.8$ & 71.6 $\pm$ $0.9$ & 43.3 $\pm$ $1.9$ & 54.6 $\pm$ $1.0$ & 60.7 $\pm$ $0.8$ \\     
    \midrule
    \redt{$\Phi_{S_1}$, $\Phi_{S_2}$} & \redt{$\theta$} & \redt{5} & \redt{ResNet-12 (pre)} & \redt{60.2 $\pm$ $1.8$} & \redt{74.3 $\pm$ $0.9$} & \redt{43.6 $\pm$ $1.8$} & \redt{55.4 $\pm$ $0.9$} & \redt{62.4 $\pm$ $0.8$} \\
  \bottomrule
\end{tabular}
  \vspace{0.2cm}
  \caption{Additional ablative study. On the last row, we show the red numbers which are reported in our main paper (corresponding to the MTL setting: \emph{SS} $[\Theta; \theta]$, meta-batch).}
    \label{table}
\end{table*}

\section{Interpretation of meta-learned \emph{SS}}
\label{sec_visual}
\begin{figure*}
\centering
\includegraphics[height=3in]{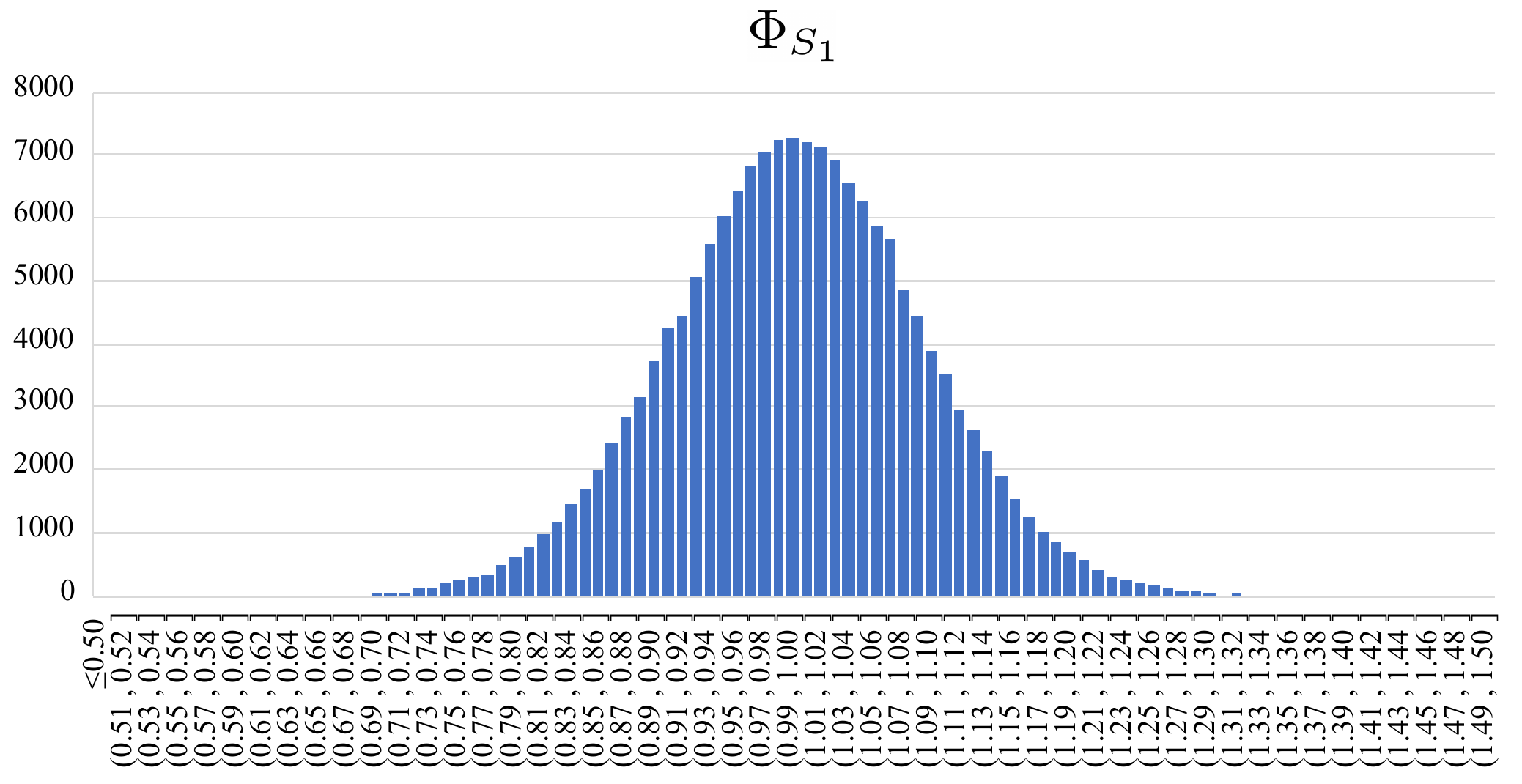}
\includegraphics[height=3in]{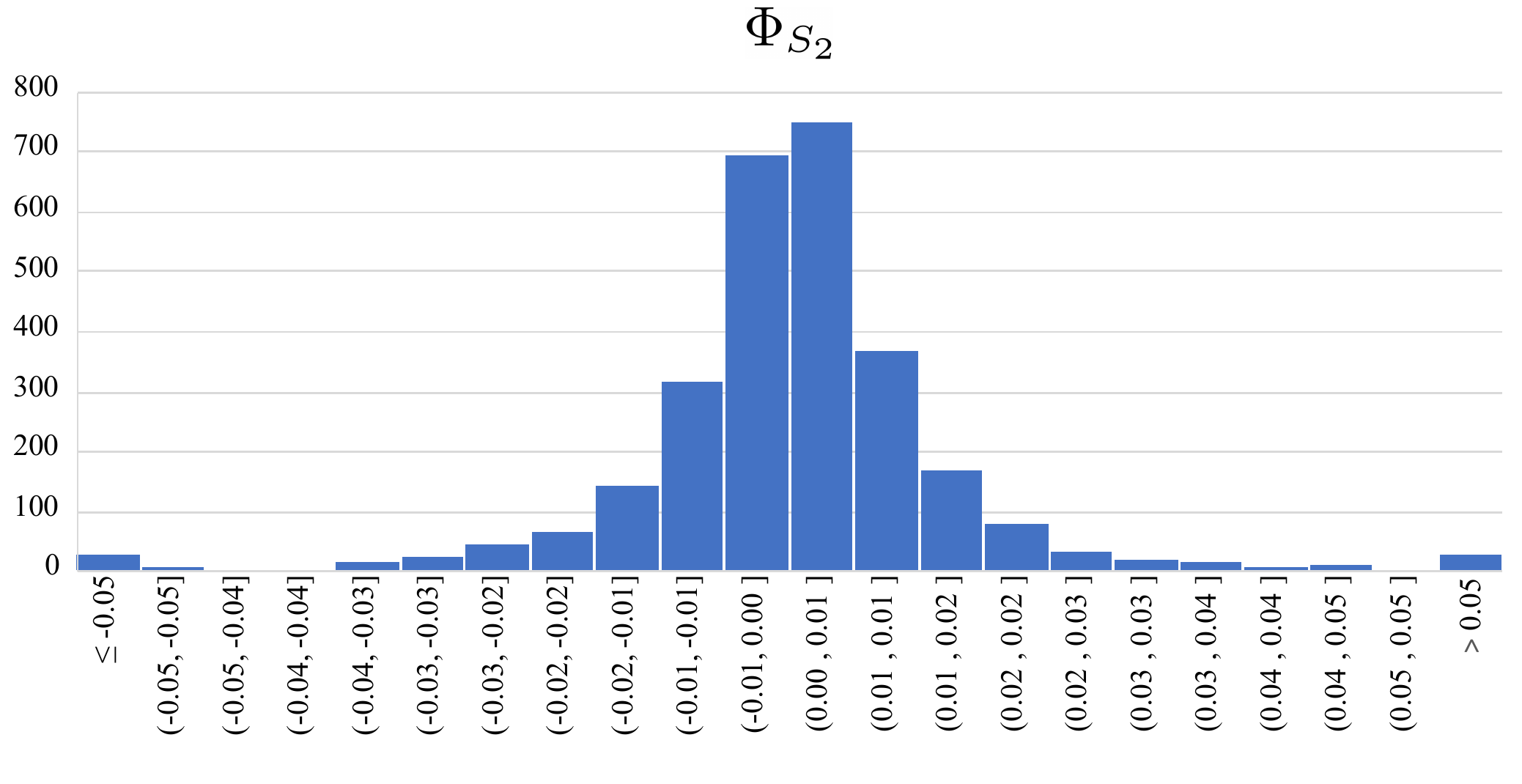}
\caption{The statistic histograms of learned \emph{SS} parameters, taking miniImageNet 1-shot as an example setting.}
\vspace{-0.3cm}
\label{statistics}
\end{figure*}

In Figure~\ref{statistics}, we show the statistic histograms of learned \emph{SS} parameters, taking miniImageNet 1-shot as an example setting.
Scaling parameters $\Phi_{S_1}$ are initialized as 1 and shifting parameters $\Phi_{S_1}$ as 0. After meta-train, we observe that these statistics are close to Gaussian distributions respectively with ($0.9962$, $0.0084$) and ($0.0003$, $0.0002$) as (mean, variance) values, which shows that the uniform initialization has been changed to Gaussian distribution through few-shot learning. 
Possible interpretations are in three-fold: 1) majority patterns trained by a large number of few-shot tasks are close to the ones trained by large-scale data; 2) tail patterns with clear scale and shift values are the ones really contributing to adapting the model to few-shot tasks; 3) tail patterns are of small quantity, enabling the fast learning convergence.

\end{document}